\documentclass[letterpaper, 10 pt, conference]{ieeeconf}  

\IEEEoverridecommandlockouts                              

\usepackage{algorithm}
\usepackage{algorithmic}
\usepackage{pifont}

\usepackage{amsmath}
\usepackage{amssymb}
\usepackage{graphicx}
\usepackage{booktabs}
\usepackage{multirow}
\usepackage{colortbl}
\usepackage{pifont}
\usepackage{xcolor}
\usepackage[
  colorlinks=true,
  linkcolor=cyan,
  urlcolor=cyan,
  citecolor=cyan
]{hyperref}
\title{\LARGE \bf
ImagiDrive: A Unified Imagination-and-Planning Framework for Autonomous Driving
}

\author {
    Jingyu Li$^{1,2*}$,
    Bozhou Zhang$^{1*}$,
    Xin Jin$^{3}$,
    Jiankang Deng$^{4}$,
    Xiatian Zhu$^{5}$,
    Li Zhang$^{1,2\dagger}$
\thanks{$\dagger$ Corresponding author, $*$ Equal contribution}%
\thanks{$^{1}$ The authors are with School of Data Science, Fudan University.}%
\thanks{$^{2}$ The authors are with Shanghai Innovation Institute.}%
\thanks{$^{3}$ The author is with Eastern Institute of Technology.}%
\thanks{$^{4}$ The author is with Imperial College London.}%
\thanks{$^{5}$ The author is with University of Surrey.}%
}

\begin{document}

\maketitle
\thispagestyle{empty}
\pagestyle{empty}

\begin{abstract}
Autonomous driving requires rich contextual comprehension and precise predictive reasoning to navigate dynamic and complex environments safely. Vision-Language Models (VLMs) and Driving World Models (DWMs) have independently emerged as powerful recipes addressing different aspects of this challenge. VLMs provide interpretability and robust action prediction through their ability to understand multi-modal context, while DWMs excel in generating detailed and plausible future driving scenarios essential for proactive planning.
Integrating VLMs with DWMs is an intuitive, promising, yet understudied strategy to exploit the complementary strengths of accurate behavioral prediction and realistic scene generation. Nevertheless, this integration presents notable challenges, particularly in effectively connecting action-level decisions with high-fidelity pixel-level predictions and maintaining computational efficiency.
In this paper, we propose {\bf \em ImagiDrive}, a novel end-to-end autonomous driving framework that integrates a VLM-based driving agent with a DWM-based scene imaginer to form a unified imagination-and-planning loop. The driving agent predicts initial driving trajectories based on multi-modal inputs, guiding the scene imaginer to generate corresponding future scenarios. These imagined scenarios are subsequently utilized to iteratively refine the driving agent’s planning decisions.
To address efficiency and predictive accuracy challenges inherent in this integration, we introduce an early stopping mechanism and a trajectory selection strategy. Extensive experimental validation on the nuScenes and NAVSIM datasets demonstrates the robustness and superiority of \textit{ImagiDrive} over previous alternatives under both open-loop and closed-loop conditions.
Code will be available at \url{https://github.com/fudan-zvg/ImagiDrive}
\end{abstract}

\section{Introduction}
\begin{figure*}[t]
    \centering
    \includegraphics[width=\textwidth]{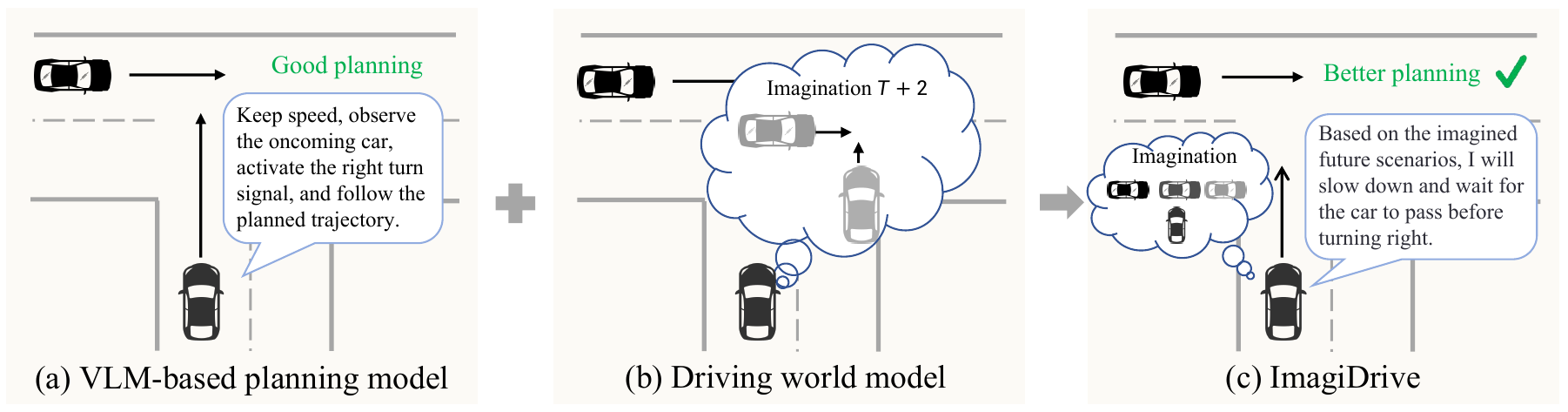}
    \caption{
    Overview of autonomous driving paradigms.
\textbf{(a)} VLM-based end-to-end methods may produce an effective planning strategy to avoid potential collisions.
\textbf{(b)} DWMs predict and generate future scenarios (T+2 seconds) to identify potential hazards.
\textbf{(c)} Our proposed framework, ImagiDrive, integrates both paradigms: Using future scene imagination from the DWM-based scene imaginer to iteratively refine VLM-based policy decisions and enhance safety.
    }
    \label{fig:intro_fig}
\end{figure*}

End-to-end autonomous driving has made significant progress, with unified models~\cite{uniad,vad,sparsedrive,bridgingAD,diffusiondrive} jointly optimizing perception, prediction, and planning on large-scale datasets. Despite strong performance, these methods often lack holistic scene understanding and causal reasoning, limiting their ability to produce rational and flexible trajectories.
Recently, vision-language models~(VLMs)~\cite{EMMA,internvl,qwen2,lmdrive} and driving world models~(DWMs)~\cite{wovogen,drivedreamer,occworld,zhang2025learning,vista,zhang2025epona} have emerged as promising alternatives. VLMs, pretrained on image-text pairs, offer strong scene comprehension, logical reasoning, and zero-shot generalization, making them ideal for cognitively inspired driving (Fig.~\ref{fig:intro_fig}(a)). Meanwhile, DWMs simulate future scenarios conditioned on past observations and potential actions, enabling agents to anticipate outcomes and evaluate decisions proactively (Fig.~\ref{fig:intro_fig}(b)).

However, integrating these two paradigms remains largely unexplored. DWMs typically focus on improving generative quality in novel scenes, while their potential in action prediction is underutilized. Some methods~\cite{vista,GenAD-world_model} generate trajectories using inverse dynamics models on generated images, but this process is overly complex and lacks deep scene understanding, often resulting in suboptimal predictions.
A natural direction is to combine the complementary strengths of VLMs and DWMs: using planning to guide imagination and imagined futures to refine planning. This integration, however, poses challenges in aligning high-level reasoning with low-level generation and mitigating the slow inference speed of both components.

To this end, we introduce \textit{\bf ImagiDrive}, an end-to-end autonomous driving framework that integrates a VLM-based driving agent with a DWM-based scene imaginer in a recurrent imagination-and-planning loop (Fig.~\ref{fig:intro_fig}(c)). 
Our driving agent can be easily integrated with mainstream vision-language models such as LLaVA~\cite{liu2023llava,liu2024llavanext} and the InternVL series~\cite{internvl}, enabling multi-modal inputs and structured outputs.
Meanwhile, our scene imaginer, with its unified input-output design, is capable of generating future scene images conditioned on different driving world models~\cite{vista,zhang2025epona}.
The core steps of our imagination-and-planning loop are summarized as follows: the agent proposes an initial trajectory from the current frame, which conditions the scene imaginer to generate future scenes. These imagined frames are then fed back into the agent to iteratively refine its planning decisions.
To ensure robust and efficient inference, we maintain a trajectory buffer to store trajectories generated in each iteration, and incorporate early stopping and trajectory selection strategies based on safety and consistency.

Our \textbf{contributions} are summarized as follows:
{\bf (i)} We present \textit{ImagiDrive}, a novel recurrent-loop autonomous driving framework that tightly couples a driving agent with a scene imaginer for imagination-driven planning.
{\bf (ii)} We develop a VLM-based driving agent that supports diverse multi-modal inputs and produces structured trajectory predictions. To better integrate with our scene imaginer, we further propose the trajectory buffer with two key strategies: early stopping and trajectory selection, which together enable efficient and reliable inference.
{\bf (iii)} Comprehensive experiments on the nuScenes and NAVSIM datasets under both open-loop and closed-loop conditions demonstrate the effectiveness and adaptability of \textit{ImagiDrive}.

\section{Related work}

\textbf{VLMs for autonomous driving.}
Large Language Models (LLMs)~\cite{gpt4} and Vision-Language Models (VLMs)~\cite{zhu2023minigpt, liu2023llava} exhibit strong multi-modal reasoning capabilities. In autonomous driving, VLMs have emerged as promising human-like agents, benefiting from large-scale datasets with language annotations~\cite{nie2023reason2drive, qian2024nuscenes}. Early works~\cite{mao2023gptdriver, wen2023dilu} leverage general-purpose GPT models but struggle with domain adaptation, prompting the development of domain-specific VLMs for better generalization.
DriveLM~\cite{sima2023drivelm} introduces Chain-of-Thought reasoning from perception to planning, while DriveMM~\cite{huang2024drivemm} unifies diverse datasets to build a generalizable VLM. 
Inspired by advances in robotic VLA models~\cite{openvla}, recent driving agents like EMMA~\cite{EMMA}, SimLingo~\cite{simlingo}, and ORION~\cite{orion} integrate vision, language, and action for planning and decision-making. However, these methods often rely on complex temporal modeling. In contrast, we propose a VLM that reasons about future behavior via imagined future scenes, without requiring explicit temporal modules.

\noindent\textbf{World models in autonomous driving.}
World models have become a growing focus in autonomous driving for predicting future scene evolution from current observations. Most approaches adopt generative models, such as autoregressive Transformers (e.g., GAIA-1~\cite{gaia-1}) or diffusion models (e.g., DriveDreamer~\cite{drivedreamer}, Drive-WM~\cite{drive-WM}), to synthesize future visual representations. Recent works enhance generation with structured constraints~\cite{drivedreamer}, view consistency~\cite{magicdrive}, or large-scale pretraining for zero-shot generalization~\cite{GenAD-world_model}. However, these models often struggle with physically plausible and detail-rich predictions under complex maneuvers. Vista~\cite{vista} addresses this by introducing structure-aware losses and larger datasets for improved long-term scene fidelity. 
Alternatively, some works explore joint generation of future world states and actions~\cite{occworld,drivinggpt}, or apply world models to end-to-end autonomous driving~\cite{drivingoccwrorld,drivearena}. 
In contrast to all these methods, we treat the world model as a future scene synthesizer, conditioned on high-quality trajectories to generate more coherent future scene contexts.

\begin{figure*}[ht]
    \centering
    \includegraphics[width=\linewidth]{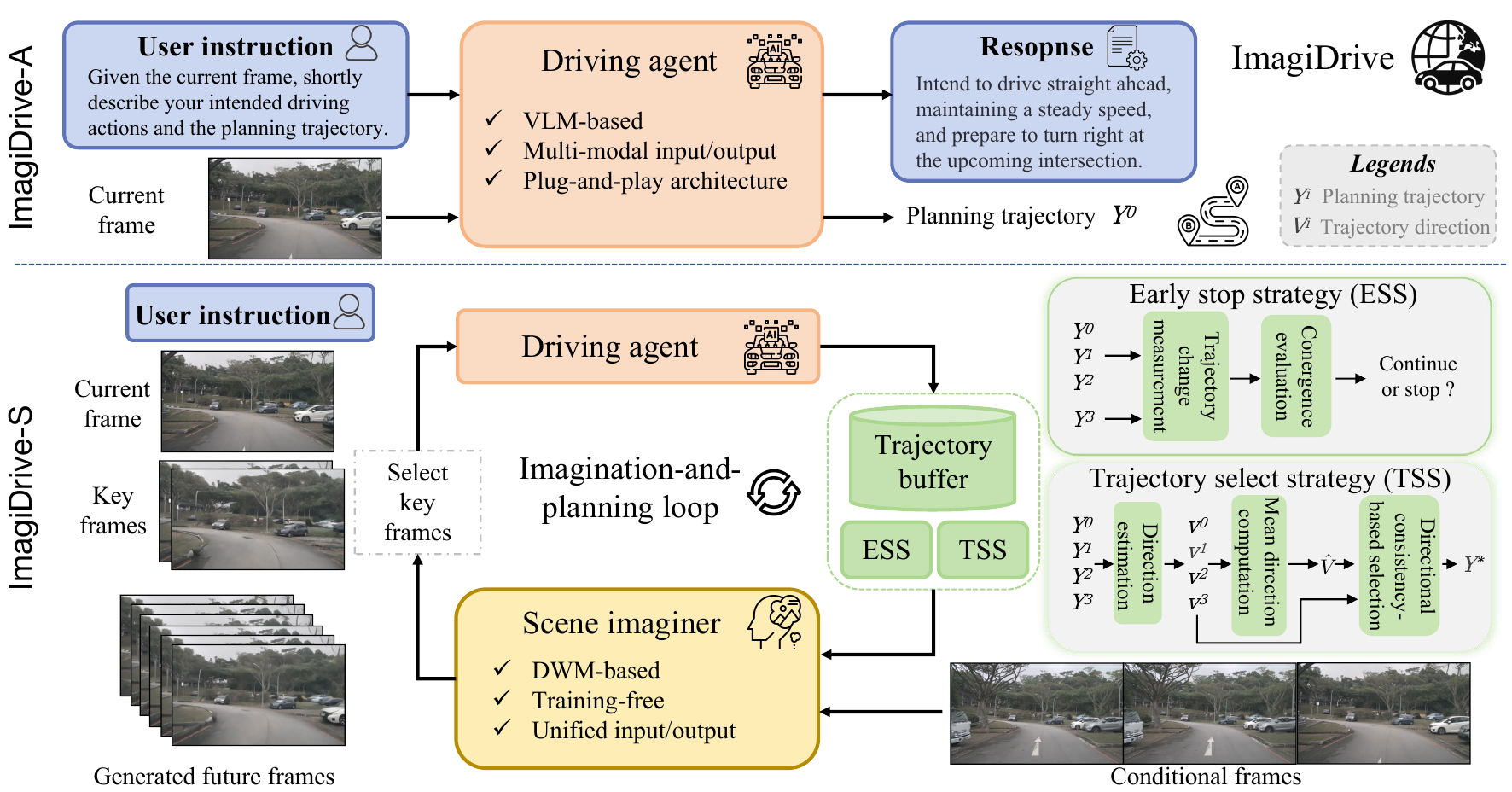}
    \caption{Overview of~\textit{ImagiDrive}. Overview of our system, which includes a driving agent, a scene imaginer, and a trajectory buffer. It operates in two modes: ImagiDrive-A is a standard planning model that uses only the driving agent, while ImagiDrive-S adopts an imagination-and-planning loop, where the scene imaginer generates future frames based on past observations and predicted trajectories. These imagined frames are iteratively fed back to refine planning. The trajectory buffer stores all trajectories, selects the best one, and decides early termination.
}
    \label{fig:pipeline}
\end{figure*}

\noindent\textbf{End-to-end autonomous driving.}
Recent advancements have shifted from isolated task pipelines to unified end-to-end frameworks that jointly address perception and ego-trajectory generation~\cite{stp3,uniad，casas2021mp3,bridgingAD}. Early works such as ST-P3~\cite{stp3} adopt intermediate representations to enable end-to-end learning, but limited scene understanding constrains planning quality. UniAD~\cite{uniad} extends this by unifying diverse subtasks, while VAD~\cite{vad} further improves modularity through vectorized representations. SparseAD~\cite{sparsead} and SparseDrive~\cite{sparsedrive} enhance efficiency and scalability via sparse inputs.
With more challenging simulators and benchmarks~\cite{navsim,Bench2Drive}, recent efforts have focused on real-world and closed-loop driving. VADv2~\cite{vadv2} achieves state-of-the-art CARLA~\cite{driveadapter,thinktwice} performance by incorporating a large trajectory vocabulary, conflict-aware loss, and probabilistic planning. DiffusionDrive~\cite{diffusiondrive} improves accuracy and diversity with a truncated diffusion policy.


\section{Method}
\textbf{Overview.}
Our \textit{ImagiDrive}, as illustrated in Fig.~\ref{fig:pipeline},
integrates a driving agent, a trajectory buffer and a scene imaginer in a {\em recurrent imagination-and-planning} framework: 
Given a current frame as input, the driving agent predicts an initial trajectory, which is then used to guide the scene imaginer in generating short-term future scene sequences;
Subsequently, selected future frames are fed back into the agent for iterative planning refinement.
To enhance efficiency and safety, we further introduce a convergence-based early stopping mechanism and a direction-consistent trajectory selection strategy.

\subsection{Driving agent} \label{methods:vlm}
\begin{figure}[t]
    \centering
    \includegraphics[width=\linewidth]{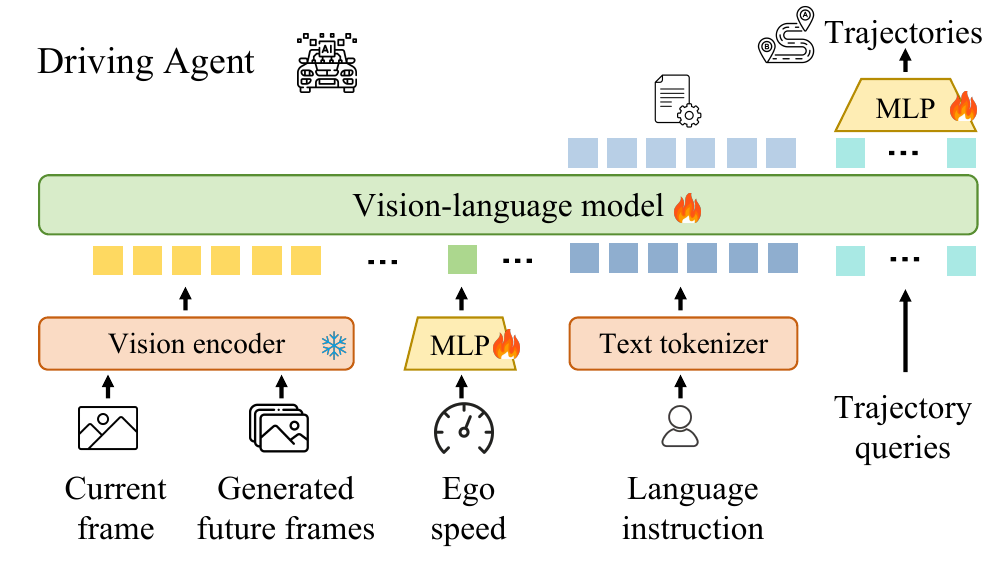}
    \caption{Overview of~\textit{Driving agent}. The agent takes multi-model inputs and produces both language and trajectory predictions. }
    \label{fig:vlm_pipeline}
\end{figure}

We build the driving agent based on the VLM with additional input modalities and output heads (Fig.~\ref{fig:vlm_pipeline}).
VLM provides a unified and simple framework that can flexibly handle single-frame or multi-frame inputs, incorporate additional driving states, and support interpretable trajectory prediction.

\noindent\textbf{Flexible input integration.}
The input of our method is multi-modal, consisting of visual data, ego state, textual prompts, and a set of trajectory queries to predict reliable driving plans.
Specifically, the visual input can be either a single front-view image or a sequence of frames that includes future predictions generated by the scene imaginer.
We encode the vehicle’s current speed as an ego state and represent it with a dedicated placeholder token~\textit{ego token}, enabling seamless integration into the VLM input stream.
Similarly, we allocate several~\textit{trajectory token}s as placeholders to indicate positions in the sequence where the model is expected to reason about the trajectory at different future time steps.
To enable adaptation to different tasks, such as predicting from the current frame or refining plans with future information, we design two prompt templates. These templates guide effective interaction among visual features, textual prompts, and trajectory tokens for plan generation.

\noindent\textbf{Vision-language model.}
Thanks to the unified token-based input and output architecture of the current VLMs, our driving agent can seamlessly support both single-frame and multi-frame inputs.
For current frame inputs, we feed the image~$I_c \in \mathbb{R}^{H \times W}$ directly into the visual encoder. In contrast, future frames $I_f \in \mathbb{R}^{H \times W}$ generated by the scene imaginer frequently exhibit artifacts such as soft edges, ghosting, and diminished texture fidelity. To address this, we apply targeted distortion augmentations during training, including Gaussian blur, shadow overlays, and random noise injection, to improve the model's robustness against such distortions.

Similar to the standard VLM processing pipeline, we use a vision encoder to process image inputs~$e_i$ and a tokenizer to handle language instruction inputs~$e_l$. In addition, we employ a lightweight MLP network to process the additional ego state information~$e_{ego}$. We initialize a set of learnable trajectory queries $q \in \mathbb{R}^{N_q \times C}$ to predict trajectory, where each query corresponds to waypoints at different future time steps, and $N_q$ denotes the number of queries.
After encoding each modality, we replace the placeholder tokens with their corresponding embeddings to form the final input sequence. The VLM then processes this sequence to produce output features:
\begin{equation}
    o_l, \tilde{q} = VLM(e_i,e_{ego},e_l,q).
\end{equation}
where $o_l$ denotes the language predictions and $\tilde{q} = \{\tilde{q}_1, \dots, \tilde{q}_{N_q}\}$ represents the contextualized trajectory queries, which are subsequently sent to the trajectory decoder to get final trajectory.

\noindent\textbf{Unified output format.}
For the language output, we perform auto-regressive decoding and apply a cross-entropy loss on the predicted tokens.
Unlike previous methods~\cite{bridgingAD,sparsedrive} that utilize multiple queries to represent diverse candidate trajectories, we design each trajectory query $\tilde{q}_i$ to correspond to a specific future time step $t_i$. We employ an MLP as the trajectory decoder and the model outputs a sequence of 2D waypoints:~$W = \{w_{t_1}, w_{t_2}, \dots, w_{t_{N_q}}\}$, where each $w_{t_i} \in \mathbb{R}^2$ denotes the predicted vehicle position at time $t_i$. This design allows the model to generate a temporally coherent trajectory by explicitly reasoning over multiple future steps.
We use Smooth L1 loss to supervise the predicted waypoints, and the overall training objective jointly optimizes the language modeling loss~$\mathcal{L}_{\text{l}}$ and the trajectory prediction loss~$\mathcal{L}_{\text{traj}}$. The total loss is shown below: 
\begin{equation}
    \mathcal{L}_{\text{total}} = \mathcal{L}_{\text{l}} + \lambda \mathcal{L}_{\text{traj}},
\end{equation}
where $\lambda$ is a weighting factor. This unified decoding framework enables the VLMs to reason holistically over both textual intent and spatiotemporal planning, producing interpretable commands and temporally structured trajectories.

\subsection{Imagination-and-planning loop}\label{methods:world_model}
Taking advantage of world models~\cite{adriver-i,diffusiondrive,infinitydrive,drivingdiffusion} in predicting future scene evolution from current observations, 
we propose to {\em directly anticipate the future scenes},
bypassing the scene dynamics modeling as done in prior works.  

\noindent\textbf{Scene imaginer.} 
Existing generative approaches for autonomous driving can be broadly categorized into diffusion-based~\cite{vista} and GPT-based methods\cite{drivinggpt,hu2024drivingworld}. 
Diffusion-based methods focus on generating future frames conditioned on reference images, and rely on future-motion-consistent frames to derive future driving plans via inverse kinematics modeling.
In contrast, GPT-based methods decouple image and motion modeling, enabling the joint generation of future visual scenes and driving plans. 
However, both paradigms face challenges in producing precise and fine-grained future trajectories due to limited capacity for comprehensive scene understanding~\cite{zhang2025epona}.
Rather than directly addressing these limitations, we leverage the strong generative capabilities of both paradigms through our Scene Imaginer: given conditional images and a continuous trajectory, it generates a temporally coherent sequence of future frames that reflect plausible driving outcomes.

\noindent\textbf{Imagination and planning.}
Initialization of our method commences by generating a preliminary trajectory $\hat{Y}_{\text{curr}}$, using the VLM-based driving agent conditioned on the current image frame.
To enhance the quality and temporal consistency of subsequent predictions by the scene imaginer, its generation process is conditioned on three preceding frames (spanning 0.25 seconds) alongside the current frame. 
Subsequently, the scene imaginer~(${SI}$) utilizes the historical image sequence $\mathbf{I}_{obs}$ and the conditional trajectory $\hat{Y}_{curr}$ to generate a short-term video prediction:
\begin{equation} \label{eq:video_prediction_mw}
    \hat{\mathbf{V}}_{fut} = SI(\mathbf{I}_{obs}, \hat{Y}_{curr})
\end{equation}
We then select two special frames from the generated future sequence as key frames, corresponding to 0.5 and 1.0 seconds into the future, respectively. These frames, together with the current frame, are used as inputs to the agent to guide subsequent trajectory prediction.

After the initial round of planning and imagination, our model proceeds iteratively: the VLM-based driving agent refines its trajectory by incorporating the current frame and selected future key frames generated by the scene imaginer, shown in Fig.~\ref{fig:pipeline}.
We store each trajectory predicted by the agent into the trajectory buffer.
This process continues for a fixed number of iterations or terminates early if a stopping criterion is triggered.
\begin{table*}[t]
\caption{\textbf{Closed-loop planning} results on NeuroNCAP~\cite{neuroncap}. ${\dagger}$ Corresponds to the model's original setting. Avg and Stat represent average and stationary, respectively. ``-'': Unreported results.}
\vspace{-10pt}
\centering
\begin{tabular}{l|cccc|cccc}
\toprule[1.5pt]
\multirow{2}{*}{\textbf{Method}} & \multicolumn{4}{c|}{\textbf{NeuroNCAP score} $\uparrow$} & \multicolumn{4}{c}{\textbf{Collision rate (\%) }$\downarrow$}   \\
& Avg. & Stat. & Frontal & Side  & Avg. & Stat. & Frontal  & Side   \\ \hline
UniAD~\cite{uniad}   & 0.73 & 0.84 & 0.10 & 1.26 & 88.60 & 87.80 & 98.40 & 79.60 \\
VAD$^\dagger~$\cite{vad}  & 0.66 & 0.47 & 0.04 & 1.45 & 92.50 & 96.20 & 99.60 & 81.60 \\ 
SparseDrive$^\dagger~$\cite{sparsedrive} &0.92  & 1.16 & 0.63 & 0.96 & 93.90 & 93.40 & 98.40 & 90.00 \\
BridgeAD~\cite{bridgingAD} & 1.60 & - & - & - & 72.60 & - & - & - \\
Impromptu VLA~\cite{ImpromptuVLA} & 2.15 & 1.77 & 2.31 & 1.75 & 65.50 & 70.00 & \textbf{59.00} & 65.00 \\
\midrule
LLava-1.6 + ImagiDrive-A (Ours)  & 2.37 & 2.89 & 1.40 & 2.81 & 66.79 & 50.13& 90.06 & 60.18 \\
LLava-1.6 + ImagiDrive-S (Ours)   & 2.99 & 3.63 & 1.88 & 3.47 & 59.45 & 45.72 & 82.15 & 50.49 \\
InternVL2.5 + ImagiDrive-A (Ours)  & 3.11 & 3.81 & 1.97 & 3.54 & 48.57 & 37.24 & 78.47 & 30.01 \\
InternVL2.5 + ImagiDrive-S (Ours) & \textbf{3.49} & \textbf{4.15} & \textbf{2.45} & \textbf{3.88} & \textbf{44.90} & \textbf{33.80} & 74.00 & \textbf{26.80} \\
\bottomrule[1.5pt]
\end{tabular}
\vspace{-10pt}
\label{tab:neuro-ncap}
\end{table*}

\noindent\textbf{Early stop strategy.}
As driving agent and scene imaginer both are inefficient,
minimizing the computation overhead is crucial for practical deployment. {\em We observe that when the predicted trajectories converge across consecutive iterations, the corresponding generated future images would become increasingly similar, yielding diminishing returns}. Motivated by this, we design an early stopping metric named {\em Trajectory Convergence Ratio}~(TCR), which enables us to adaptively terminate the iterative process once the trajectory change diminishes.
Specifically, TCR captures the normalized rate of change across different time steps of a trajectory, enabling stable and consistent evaluation regardless of the trajectory’s scale or magnitude, formulated as:
\begin{equation}
    \text{TCR}(\hat{Y}^{(i)}, \hat{Y}^{(j)}) = \frac{1}{N_q} \sum_{t=1}^{N_q} \frac{ \left\| \hat{Y}^{(i)}_t - \hat{Y}^{(j)}_t \right\|_2 }{ \left\| \hat{Y}^{(j)}_t \right\|_2 + \varepsilon },
\end{equation}
where $\hat{Y}^{(0)}$ and $\hat{Y}^{(i)}$ denote the initial and the predicted trajectories at iteration $i$, respectively. $N_q$ is the number of predicted waypoints and $\varepsilon$ is a small constant added for numerical stability to avoid division by zero. At each iteration, we compute the TCR between $\hat{Y}^{(i)}$ and all previous predictions $\hat{Y}^{(j)}$ $(j < i)$. The iterative process will terminate once the TCR falls below a predefined threshold~$\theta$, indicating sufficient convergence.

\noindent\textbf{Trajectory selection strategy.}
To obtain more robust trajectories, we design a trajectory selection strategy based on the concept of modal directionality from directional statistics, emphasizing trend direction consistency. 
Given a set of predicted trajectories $\{\hat{Y}^{(i)}\}_{i=1}^{N}$, we first compute the unit average direction vector for each trajectory:
\begin{equation}
    \hat{\mathbf{v}}^{(i)} = \frac{1}{n-1} \sum_{t=1}^{n-1} \frac{\mathbf{p}^{(i)}_{t+1} - \mathbf{p}^{(i)}_t}{\|\mathbf{p}^{(i)}_{t+1} - \mathbf{p}^{(i)}_t\| + \varepsilon},
\end{equation}
where $p_t$ is the waypoints at time $t$. The mean direction vector across all predictions is:
\begin{equation}
    \hat{\bar{\mathbf{v}}} = \frac{1}{N} \sum_{i=1}^{N} \hat{\mathbf{v}}^{(i)}, \quad \hat{\bar{\mathbf{v}}} \leftarrow \frac{\hat{\bar{\mathbf{v}}}}{\|\hat{\bar{\mathbf{v}}}\| + \varepsilon}.
\end{equation}

We then compute the angle between each predicted direction and the global mean. The final selected trajectory is the one {\em most aligned} with the average direction:
\begin{align}
    \theta^{(i)} &= \arccos\left( \mathrm{clip}\left(\hat{\mathbf{v}}^{(i)} \cdot \hat{\bar{\mathbf{v}}}, -1, 1\right) \right), \\
    i^* &= \arg\min_i \theta^{(i)}, \quad \hat{Y}^* = \hat{Y}^{(i^*)}.
\end{align}
\begin{table*}[t] 
 \caption{\textbf{Open-loop planning} results on the Turning-nuScenes validation dataset~\cite{momAD}. We follow the ST-P3~\cite{stp3} evaluation metric.}
 \vspace{-10pt}
        \centering
        \begin{tabular}{l|cccc|cccc}
        \toprule[1.5pt]
        \multirow{2}{*}{\textbf{Method}} &
        \multicolumn{4}{c|}{\textbf{L2 ($m$)} $\downarrow$} & 
        \multicolumn{4}{c}{\textbf{Col. Rate (\%)} $\downarrow$} \\
         & 1$s$ & 2$s$ & 3$s$ & Avg. & 1$s$ & 2$s$ & 3$s$ & Avg.\\
        \midrule 
         UniAD~\cite{uniad} & 0.52 & 0.88 & 1.64  & 1.01 & 0.16 & 0.51 & 1.41 & 0.69 \\
         VAD~\cite{vad}  & 0.48 & 0.80 & 1.55 & 0.94 & 0.07 & 0.41 & 1.20 & 0.56\\
         SparseDrive~\cite{sparsedrive}  & 0.35 & 0.77 & 1.46 & 0.86 & 0.04 & 0.17 & 0.98 & 0.40\\
         MomAD~\cite{momAD} & \textbf{0.33} & 0.70 & 1.24 & 0.76 & 0.03 & 0.13 & 0.79 & 0.32\\
         \midrule
         LLava-1.6 + ImagiDrive-A (Ours) & 0.62 &  1.06 &  1.71 &  1.13 &  0.17 &  0.41 &  1.37 &  0.65\\
         LLava-1.6 + ImagiDrive-S (Ours) & 0.48 & 0.82 & 1.50 & 0.93 & 0.09 & 0.32 & 1.08 &  0.50\\
         InternVL2.5 + ImagiDrive-A (Ours) &   0.40 &  0.81 &  1.35 &  0.85 &  \textbf{0.00} &  0.28 &  0.89 &  0.39\\
         InternVL2.5 + ImagiDrive-S (Ours)  & 0.34 & \textbf{0.69} & \textbf{1.23} & \textbf{0.75} & \textbf{0.00} & \textbf{0.12} & \textbf{0.53} & \textbf{0.22} \\
         \bottomrule[1.5pt]
        \end{tabular}
   \vspace{-10pt}
\label{tab:openloop_turn}
\end{table*}
\section{Experiments}
\subsection{Datasets}
We evaluate our method on three datasets, including both closed-loop and open-loop settings.
NeuroNCAP~\cite{neuroncap} simulator, a photorealistic platform built on nuScenes that enables safety-critical closed-loop evaluation.
Turning-nuScenes~\cite{momAD}, a smaller yet more challenging dataset. It contains a challenging subset of turning scenes, including 17 scenes with 680 samples. 
NAVSIM~\cite{navsim} is a large-scale real-world autonomous driving dataset designed for non-reactive simulation and benchmarking.
It is built upon OpenScene~\cite{openscene2023} and focuses on challenging scenarios involving dynamic intention changes, while filtering out trivial cases such as stationary scenes or constant-speed driving.

\subsection{Implementation details}
Since our driving agent is easily integrated into VLMs, we select two representative VLM models, LLaVA-1.6-7B~\cite{liu2024llavanext} and InternVL2.5-4B~\cite{internvl}, to serve as our driving agents. Meanwhile, we adopt Vista~\cite{vista} as the scene imaginer on the nuScenes dataset and Epona~\cite{zhang2025epona} as the scene imaginer on the NAVSIM dataset.

For our driving agent, we first train our driving agent with OmniDrive~\cite{omnidrive} for 3 epochs, facilitating alignment between vision and language in autonomous driving scenarios. We then design a planning QA template with special trajectory tokens and train the model for 5 epochs. We use data augmentation techniques for future frames to simulate the distortion caused by the diffusion model during training. 
We set the number of trajectory queries as 6, $\lambda_1$ and $\lambda_2$ as 0.1 and 0.5, respectively. The model is trained on 4 NVIDIA H100 GPUs with a batch size of 1 for 12 hours.

For our Imagination-and-planning loop, we set $\theta$ as 0.05 empirically, and the maximum of iterations for the recurrent loop as 5. 
Our \textit{ImagiDrive} has two variants: {\bf ImagiDrive-A}, a VLM-based agent relying solely on the current frame, and \textbf{ImagiDrive-S}, with a scene imaginer to generate and incorporate two future frames, enabling more informed planning.

\begin{table*}[t!]
    \centering
    \caption{\textbf{Planning performance on the NAVSIM~\cite{navsim} test set evaluated with closed-loop metrics.} NC: no at-fault collision; DAC: drivable area compliance; TTC: time-to-collision; Comf.: comfort; EP: ego progress; PDMS: predictive driver model score. Our world model outperforms strong end-to-end planners in terms of overall PDMS. The upper section presents end-to-end methods trained with dense scene annotations, while the lower section includes methods that are assisted by a world model without relying on dense supervision.}
    \vspace{-10pt}
    \setlength{\tabcolsep}{1.9mm}
    \begin{tabular}{lc|ccccc|c}
        \toprule[1.5pt]
        \textbf{Method} & \textbf{Input} & \textbf{NC} $\uparrow$ & \textbf{DAC} $\uparrow$ & \textbf{TTC} $\uparrow$ & \textbf{Comf.} $\uparrow$ & \textbf{EP} $\uparrow$ & \textbf{PDMS} $\uparrow$ \\
        \midrule
        VADv2~\cite{vadv2} & Camera \& Lidar & 97.2 & 89.1 & 91.9 & 100 & 76.0 & 80.9 \\
        TransFuser~\cite{Transfuser} & Camera \& Lidar & 97.7 & 92.8 & 92.8 & 100 & 79.2 & 84.0 \\
        UniAD~\cite{uniad} & Camera & 97.8 & 91.9 & 92.9 & 100 & 78.8 & 83.4 \\
        PARA-Drive~\cite{para-drive} & Camera & 97.9 & 92.4 & 93.0 & 99.8 & 79.3 & 84.6 \\
        DRAMA~\cite{drama} & Camera \& Lidar & 98.0 & 93.1 & 94.8 & 100 & 80.1 & 85.5 \\
        \midrule
        DrivingGPT~\cite{drivinggpt} & Camera & \bf 98.9 & 90.7 & \bf 94.9 & 95.6 & 79.7 & 82.4 \\
        World4Drive~\cite{world4drive} & Camera & 97.4 & 94.3 & 92.8 & \bf 100.0 & 79.9 & 85.1 \\
        Epona~\cite{zhang2025epona} & Camera & 97.9 & 95.1 & 93.8 & 99.9& 80.4 & 86.2 \\
        LLava-1.6 + ImagiDrive-A (Ours) & Camera & 97.7 & 95.3 & 93.0 & 99.9 & 80.4 & 86.0\\
         LLava-1.6 + ImagiDrive-S (Ours) & Camera & 97.9 & 95.5 & 93.1 & 99.9 & \bf 80.7 & 86.4 \\
         InternVL2.5 + ImagiDrive-A (Ours) & Camera & 98.1 & \bf 96.2 & 94.4 & \bf 100.0 & 80.1 & 86.9 \\
         InternVL2.5 + ImagiDrive-S (Ours)  & Camera & 98.6 & \bf 96.2 & 94.5 & \bf 100.0 & 80.5 & \bf 87.4 \\
        \bottomrule[1.5pt]
    \end{tabular}
    \vspace{-10pt}
    \label{tab:plan_nasim}
\end{table*}

\subsection{Results analysis}
\noindent\textbf{Results on NeuroNCAP.}
We adopt NeuroNCAP~\cite{neuroncap} as the closed-loop evaluation benchmark. 
As shown in Table~\ref{tab:neuro-ncap}, regardless of whether LLaVA-1.6 or InternVL 2.5 is used, our ImagiDrive-A consistently and significantly outperforms the existing state-of-the-art E2E method SparseDrive~\cite{sparsedrive}. 
Our method significantly outperforms ImpromptuVLA~\cite{ImpromptuVLA}, a model trained on extensive additional data, across all metrics with the sole exception of a slightly higher Frontal collision rate.
This effectively demonstrates that our agent is capable of understanding the current scene and taking appropriate actions to avoid potential hazards.
When using ImagiDrive-S, thanks to our scene imaginer and the imagination-and-planning loop, our method achieves improvements of 0.62 and 0.38 in NeuroNCAP score, along with collision rate reductions of 7.34\% and 3.67\%, respectively.
The improved performance of ImagiDrive-S compared to ImagiDrive-A indicates that incorporating generated future frames as input effectively helps the agent anticipate and avoid potential hazards, fulfilling the goal of integrating imagination and planning in a unified framework.

\noindent\textbf{Results on Turning-nuScenes.}
We conduct open-loop evaluation on the Turning-nuScenes dataset~\cite{momAD}. Turning-nuScenes is a subset of nuScenes that focuses on turning scenarios. Since the majority of scenes in nuScenes involve straight driving, the turning subset presents greater challenges for planning.
A combination of LLaVA-1.6~\cite{liu2024llavanext} and ImagiDrive-A slightly underperforms UniAD~\cite{uniad}. With the integration of our scene imaginer, the performance exceeds that of VAD~\cite{vad}, suggesting that incorporating imagined future scenes introduces richer contextual information, leading to more accurate trajectory prediction and notably lower collision rates.
Benefiting from the strong performance of InternVL~\cite{internvl}, our InternVL2.5-ImagiDrive-A achieves low collision rates and trajectory deviation errors. Moreover, by introducing the imagination-and-planning loop, InternVL2.5-ImagiDrive-S further improves the performance, surpassing MomAD and significantly reducing the collision rate. 
The results under challenging scenarios further demonstrate that our proposed imagination-and-planning framework effectively leverages the strengths of both the VLM and DWM, leading to more robust and efficient performance.

\noindent\textbf{Results on NAVSIM.}
We conduct experiments on the NAVSIM~\cite{navsim} dataset with closed-loop metrics, as shown in Table~\ref{tab:plan_nasim}. We compare our ImagiDrive with some end-to-end methods~\cite{vadv2,uniad}. Leveraging the VLM's superior scene understanding capabilities, ImagiDrive-A significantly outperforms end-to-end methods on PDMS. Furthermore, ImagiDrive-S fully exploits the generative capability of the scene imaginer and the reasoning strength of the VLM through the imagination-and-planning loop, leading to superior performance on the EP edge cases compared to purely world model-based approaches~\cite{zhang2025epona,world4drive}.

\begin{figure*}[h!]
    \centering
    \includegraphics[width=\linewidth]{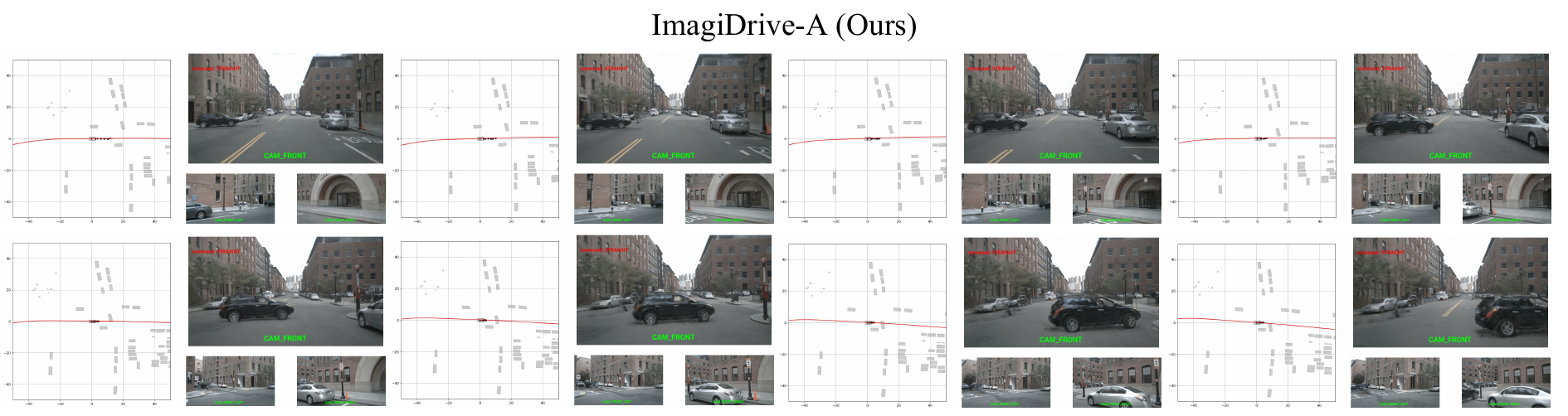}
    \caption{Qualitative results in the closed-loop evaluation demonstrate that our ImagiDrive effectively avoids collisions in intersection side-encounter scenario.}
    \label{fig:vis_closed}
    \includegraphics[width=\linewidth]{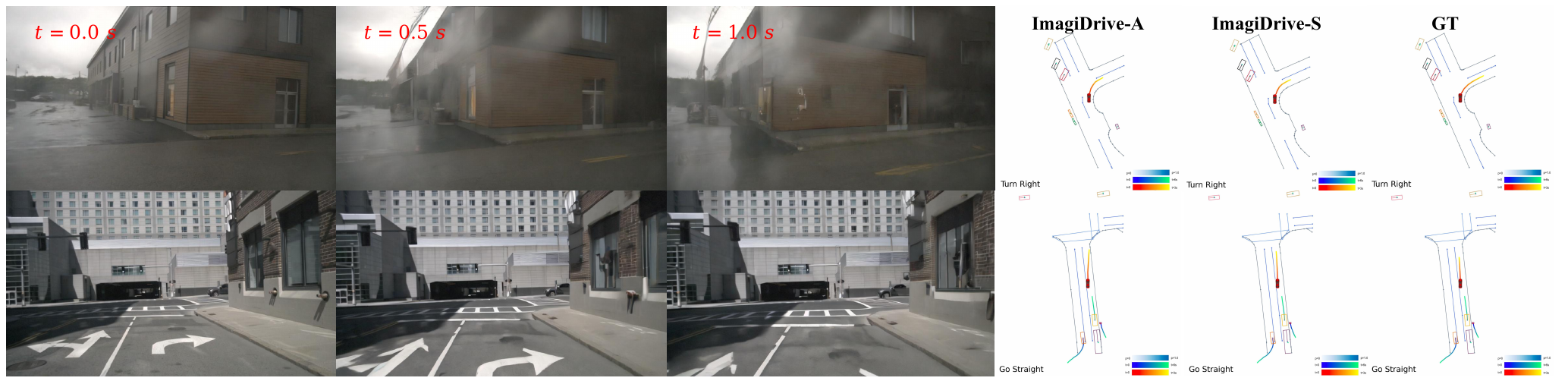}
    \caption{Qualitative results from the open-loop evaluation show that our ImagiDrive can effectively correct the trajectory.}
    \label{fig:vis_open}
\end{figure*}

\begin{table}[t]
 \caption{Ablations on early stop strategy~(ESS) and trajectory selection stretegy~(TSS).}
 \vspace{-10pt}
 \centering
    \setlength{\tabcolsep}{1.mm}
    \begin{tabular}{cc|cc}
        \toprule[1.5pt]
        \textbf{ESS} & \textbf{TSS}&
        \textbf{Avg. Col. Rate (\%)} $\downarrow$ &
        \textbf{Avg. Iterations~(steps)}  $\downarrow$\\
        \midrule 
         &  &  0.39 & 5\\
         \ding{51} & & 0.28&2.3\\
         & \ding{51} & 0.22 &5\\
         \ding{51}& \ding{51}&0.21&2.3\\
         \bottomrule[1.5pt]
         \end{tabular}
        
         \label{tab:ablation}
         \vspace{-10pt}
\end{table}
\begin{table}[t] 
        \centering
        \caption{Ablation Study on Different Trajectory Selection Strategies.}
        \vspace{-10pt}
        \begin{tabular}{c|cccc}
        \toprule[1.5pt]
        \multirow{2}{*}{Method} &
        \multicolumn{4}{c}{\textbf{Col. Rate (\%)} $\downarrow$} \\
        & 2.0$s$ & 2.5 $s$ & 3.0$s$& \textit{Avg.} \\
        \midrule 
        SmoothSel& 0.18& 0.37& 0.58& 0.38\\
        SoftMin& 0.23 &  0.44& 0.67& 0.45\\
        MaxCons& 0.18 & 0.37 & 0.61& 0.39\\
        Ours& \textbf{0.14}& \textbf{0.33}& \textbf{0.55}& \textbf{0.34} \\
         \bottomrule[1.5pt]
        \end{tabular}
    \vspace{-10pt}
    \label{tab:abla_tss}
\end{table}
\subsection{Ablation study}
\noindent\textbf{Main ablation study}. We conduct ablation studies to effect of early stop strategy and trajectory selection strategy on the Turning-nuScenes~\cite{momAD}. As shown in Table~\ref{tab:ablation}, 
We report the results in terms of collision rate and iteration steps per trajectory. Without either strategy, the average collision rate is 0.39\%, and the average number of iterations per trajectory is 5.
While applying ESS alone reduces inference time significantly, it slightly increases collision rate. In contrast, TSS alone improves collision avoidance without affecting runtime. Combining both strategies leads to a favorable trade-off: achieving low collision rate while nearly halving the number of iterations. This validates the effectiveness of our design in balancing safety and efficiency.

\noindent\textbf{Ablation study on Trajectory Selection Strategies.} 
We compare our trajectory selection strategy on Turning-nuScene~\cite{momAD} with other strategies and the results are shown in Table~\ref{tab:abla_tss}. 1) Smoothness-Based Selection~(MaxCons), this method selects the trajectory with the lowest curvature variation, encouraging smooth and continuous motion patterns. 2) Soft-Min Weighted Averaging~(SoftMin), this approach performs a soft-min weighted average over all candidate trajectories based on a predefined cost metric. 3) Trajectory with Maximum Directional Consistency~(MaxCons), this strategy selects the trajectory that best aligns with the historical motion direction, favoring smoother transitions. Our method achieves the lowest collision rates across all time horizons, with an average of 0.34\%, outperforming all baseline strategies. While SmoothSel and MaxCons offer relatively stable performance, SoftMin lags behind. This highlights the effectiveness of our selection strategy in enhancing safety under long-term prediction.

\subsection{Qualitative analysis}
We illustrate the closed-loop results for a safety-critical scenario. As shown in Fig~\ref{fig:vis_closed}, our ImagiDrive is capable of timely deceleration to avoid obstacles upon detecting other vehicles, and gradually accelerates back to normal speed after completing the passing maneuver. 
We also present qualitative results in open-loop scenarios, as shown in Fig~\ref{fig:vis_open}, ImagiDrive-A exhibits a significant deviation from the ground-truth trajectory when predicting a right turn, which may lead to a potential collision. In contrast, ImagiDrive-S, with the aid of the world model to generate future visual context, effectively corrects the trajectory.

\section{Conclusion}
We propose \textit{ImagiDrive}, a novel end-to-end autonomous driving framework that unifies a driving agent and a scene imaginer into an imagination-and-planning paradigm. The system forms a recurrent loop: it predicts trajectories from a single image, generates future images conditioned on the trajectory, and refines the trajectory using these imagined frames. To enhance efficiency, we introduce an early stopping strategy and a trajectory selection strategy. 
ImagiDrive achieves compelling results in closed-loop evaluations and outperforms previous methods on the challenging Turning-nuScenes and NAVSIM dataset, reflecting its strength in complex scenarios. 

\section{Acknowledgements} 
This work was supported in part by New Generation Artificial Intelligence-National Science and Technology Major
Project (2025ZD0123004), Ningbo grant (2025Z038) and National Natural Science Foundation of China (Grant No. 62376060).

{\small
\bibliographystyle{IEEEtran}
\bibliography{IEEEabrv}

@String(CVPR= {IEEE Conf. Comput. Vis. Pattern Recog.})

@String(ICCV= {Int. Conf. Comput. Vis.})

@String(ECCV= {Eur. Conf. Comput. Vis.})

@String(NIPS= {Adv. Neural Inform. Process. Syst.})

@String(ICLR = {Int. Conf. Learn. Represent.})

@String(AAAI = {AAAI})

@String(CVPR  = {CVPR})

@String(ICCV  = {ICCV})

@String(ECCV  = {ECCV})

@String(NIPS  = {NeurIPS})

@String(ICLR  = {ICLR})

@String(ICRA = {ICRA})

@String(NeurIPS  = {NeurIPS})

@inproceedings{uniad,
  title={Planning-oriented autonomous driving},
  author={Hu, Yihan and Yang, Jiazhi and Chen, Li and Li, Keyu and Sima, Chonghao and Zhu, Xizhou and Chai, Siqi and Du, Senyao and Lin, Tianwei and Wang, Wenhai and others},
  booktitle=CVPR,
  year={2023}
}

@inproceedings{vad,
  title={Vad: Vectorized scene representation for efficient autonomous driving},
  author={Jiang, Bo and Chen, Shaoyu and Xu, Qing and Liao, Bencheng and Chen, Jiajie and Zhou, Helong and Zhang, Qian and Liu, Wenyu and Huang, Chang and Wang, Xinggang},
  booktitle=ICCV,
  year={2023}
}

@inproceedings{sparsedrive,
  title={Sparsedrive: End-to-end autonomous driving via sparse scene representation},
  author={Sun, Wenchao and Lin, Xuewu and Shi, Yining and Zhang, Chuang and Wu, Haoran and Zheng, Sifa},
  booktitle=ICRA,
  year={2025}
}

@inproceedings{bridgingAD,
  title={Bridging Past and Future: End-to-End Autonomous Driving with Historical Prediction and Planning},
  author={Zhang, Bozhou and Song, Nan and Jin, Xin and Zhang, Li},
  booktitle=CVPR,
  year={2025}
}

@article{EMMA,
  title={Emma: End-to-end multimodal model for autonomous driving},
  author={Hwang, Jyh-Jing and Xu, Runsheng and Lin, Hubert and Hung, Wei-Chih and Ji, Jingwei and Choi, Kristy and Huang, Di and He, Tong and Covington, Paul and Sapp, Benjamin and others},
  journal={arXiv preprint arXiv:2410.23262},
  year={2024}
}

@inproceedings{lmdrive,
  title={Lmdrive: Closed-loop end-to-end driving with large language models},
  author={Shao, Hao and Hu, Yuxuan and Wang, Letian and Song, Guanglu and Waslander, Steven L and Liu, Yu and Li, Hongsheng},
  booktitle=CVPR,
  year={2024}
}

@inproceedings{internvl,
  title={Internvl: Scaling up vision foundation models and aligning for generic visual-linguistic tasks},
  author={Chen, Zhe and Wu, Jiannan and Wang, Wenhai and Su, Weijie and Chen, Guo and Xing, Sen and Zhong, Muyan and Zhang, Qinglong and Zhu, Xizhou and Lu, Lewei and others},
  booktitle=CVPR,
  year={2024}
}

@article{qwen2,
  title={Qwen2-vl: Enhancing vision-language model's perception of the world at any resolution},
  author={Wang, Peng and Bai, Shuai and Tan, Sinan and Wang, Shijie and Fan, Zhihao and Bai, Jinze and Chen, Keqin and Liu, Xuejing and Wang, Jialin and Ge, Wenbin and others},
  journal={arXiv preprint arXiv:2409.12191},
  year={2024}
}

@inproceedings{wovogen,
  title={Wovogen: World volume-aware diffusion for controllable multi-camera driving scene generation},
  author={Lu, Jiachen and Huang, Ze and Yang, Zeyu and Zhang, Jiahui and Zhang, Li},
  booktitle=ECCV,
  year={2024}
}

@inproceedings{drivedreamer,
  title={DriveDreamer: Towards Real-World-Drive World Models for Autonomous Driving},
  author={Wang, Xiaofeng and Zhu, Zheng and Huang, Guan and Chen, Xinze and Zhu, Jiagang and Lu, Jiwen},
  booktitle=ECCV,
  year={2024},
}

@inproceedings{drive-WM,
  title={Driving into the future: Multiview visual forecasting and planning with world model for autonomous driving},
  author={Wang, Yuqi and He, Jiawei and Fan, Lue and Li, Hongxin and Chen, Yuntao and Zhang, Zhaoxiang},
  booktitle=CVPR,
  year={2024}
}

@inproceedings{zhang2025learning,
  title={Learning unsupervised world models for autonomous driving via discrete diffusion},
  author={Zhang, Lunjun and Xiong, Yuwen and Yang, Ze and ROMERO, Sergio CASAS and Urtasun, Raquel},
  booktitle=ICLR,
  year={2024}
}

@inproceedings{occworld,
  title={Occworld: Learning a 3d occupancy world model for autonomous driving},
  author={Zheng, Wenzhao and Chen, Weiliang and Huang, Yuanhui and Zhang, Borui and Duan, Yueqi and Lu, Jiwen},
  booktitle=ECCV,
  year={2024},
}

@inproceedings{vista,
  title={Vista: A generalizable driving world model with high fidelity and versatile controllability},
  author={Gao, Shenyuan and Yang, Jiazhi and Chen, Li and Chitta, Kashyap and Qiu, Yihang and Geiger, Andreas and Zhang, Jun and Li, Hongyang},
  booktitle=NeurIPS,
  year={2024}
}

@inproceedings{omnidrive,
  title={OmniDrive: A Holistic Vision-Language Dataset for Autonomous Driving with Counterfactual Reasoning},
  author={Wang, Shihao and Yu, Zhiding and Jiang, Xiaohui and Lan, Shiyi and Shi, Min and Chang, Nadine and Kautz, Jan and Li, Ying and Alvarez, Jose M},
  booktitle=CVPR,
  year={2025}
}

@article{orion,
  title={ORION: A Holistic End-to-End Autonomous Driving Framework by Vision-Language Instructed Action Generation},
  author={Fu, Haoyu and Zhang, Diankun and Zhao, Zongchuang and Cui, Jianfeng and Liang, Dingkang and Zhang, Chong and Zhang, Dingyuan and Xie, Hongwei and Wang, Bing and Bai, Xiang},
  journal={arXiv preprint arXiv:2503.19755},
  year={2025}
}

@inproceedings{GenAD-world_model,
  title={Generalized predictive model for autonomous driving},
  author={Yang, Jiazhi and Gao, Shenyuan and Qiu, Yihang and Chen, Li and Li, Tianyu and Dai, Bo and Chitta, Kashyap and Wu, Penghao and Zeng, Jia and Luo, Ping and others},
  booktitle=CVPR,
  year={2024}
}

@article{gaia-1,
  title={Gaia-1: A generative world model for autonomous driving},
  author={Hu, Anthony and Russell, Lloyd and Yeo, Hudson and Murez, Zak and Fedoseev, George and Kendall, Alex and Shotton, Jamie and Corrado, Gianluca},
  journal={arXiv preprint arXiv:2309.17080},
  year={2023}
}

@inproceedings{simlingo,
  title={SimLingo: Vision-Only Closed-Loop Autonomous Driving with Language-Action Alignment},
  author={Renz, Katrin and Chen, Long and Arani, Elahe and Sinavski, Oleg},
  booktitle=CVPR,
  year={2025}
}

@inproceedings{diffusiondrive,
  title={DiffusionDrive: Truncated Diffusion Model for End-to-End Autonomous Driving},
  author={Bencheng Liao and Shaoyu Chen and Haoran Yin and Bo Jiang and Cheng Wang and Sixu Yan and Xinbang Zhang and Xiangyu Li and Ying Zhang and Qian Zhang and Xinggang Wang},
   booktitle=CVPR,
   year={2025},
}

@article{adriver-i,
  title={Adriver-i: A general world model for autonomous driving},
  author={Jia, Fan and Mao, Weixin and Liu, Yingfei and Zhao, Yucheng and Wen, Yuqing and Zhang, Chi and Zhang, Xiangyu and Wang, Tiancai},
  journal={arXiv preprint arXiv:2311.13549},
  year={2023}
}

@article{infinitydrive,
  title={InfinityDrive: Breaking Time Limits in Driving World Models},
  author={Guo, Xi and Ding, Chenjing and Dou, Haoxuan and Zhang, Xin and Tang, Weixuan and Wu, Wei},
  journal={arXiv preprint arXiv:2412.01522},
  year={2024}
}

@inproceedings{drivingdiffusion,
  title={DrivingDiffusion: Layout-Guided Multi-View Driving Scenarios Video Generation with Latent Diffusion Model},
  author={Li, Xiaofan and Zhang, Yifu and Ye, Xiaoqing},
  booktitle=ECCV,
  year={2025}
}

@inproceedings{stp3,
  title={St-p3: End-to-end vision-based autonomous driving via spatial-temporal feature learning},
  author={Hu, Shengchao and Chen, Li and Wu, Penghao and Li, Hongyang and Yan, Junchi and Tao, Dacheng},
  booktitle=ECCV,
  year={2022},
}

@article{sparsead,
  title={SparseAD: Sparse Query-Centric Paradigm for Efficient End-to-End Autonomous Driving},
  author={Zhang, Diankun and Wang, Guoan and Zhu, Runwen and Zhao, Jianbo and Chen, Xiwu and Zhang, Siyu and Gong, Jiahao and Zhou, Qibin and Zhang, Wenyuan and Wang, Ningzi and others},
  journal={arXiv preprint arXiv:2404.06892},
  year={2024}
}

@article{vadv2,
  title={Vadv2: End-to-end vectorized autonomous driving via probabilistic planning},
  author={Chen, Shaoyu and Jiang, Bo and Gao, Hao and Liao, Bencheng and Xu, Qing and Zhang, Qian and Huang, Chang and Liu, Wenyu and Wang, Xinggang},
  journal={arXiv preprint arXiv:2402.13243},
  year={2024}
}

@inproceedings{navsim,
  title={Navsim: Data-driven non-reactive autonomous vehicle simulation and benchmarking},
  author={Dauner, Daniel and Hallgarten, Marcel and Li, Tianyu and Weng, Xinshuo and Huang, Zhiyu and Yang, Zetong and Li, Hongyang and Gilitschenski, Igor and Ivanovic, Boris and Pavone, Marco and others},
  booktitle=NIPS,
  year={2024},
}

@inproceedings{Bench2Drive,
  title={Bench2Drive: Towards Multi-Ability Benchmarking of Closed-Loop End-To-End Autonomous Driving},
  author={Xiaosong Jia and Zhenjie Yang and Qifeng Li and Zhiyuan Zhang and Junchi Yan},
  booktitle=NeurIPS,
  year={2024}
}

@inproceedings{thinktwice,
  title={Think twice before driving: Towards scalable decoders for end-to-end autonomous driving},
  author={Jia, Xiaosong and Wu, Penghao and Chen, Li and Xie, Jiangwei and He, Conghui and Yan, Junchi and Li, Hongyang},
  booktitle=CVPR,
  year={2023}
}

@inproceedings{driveadapter,
  title={Driveadapter: Breaking the coupling barrier of perception and planning in end-to-end autonomous driving},
  author={Jia, Xiaosong and Gao, Yulu and Chen, Li and Yan, Junchi and Liu, Patrick Langechuan and Li, Hongyang},
  booktitle=ICCV,
  year={2023}
}

@inproceedings{magicdrive,
  title={MagicDrive: Street View Generation with Diverse 3D Geometry Control},
  author={Gao, Ruiyuan and Chen, Kai and Xie, Enze and Lanqing, HONG and Li, Zhenguo and Yeung, Dit-Yan and Xu, Qiang},
  booktitle=ICLR,
  year={2024}
}

@inproceedings{drivingoccwrorld,
  title={Driving in the occupancy world: Vision-centric 4d occupancy forecasting and planning via world models for autonomous driving},
  author={Yang, Yu and Mei, Jianbiao and Ma, Yukai and Du, Siliang and Chen, Wenqing and Qian, Yijie and Feng, Yuxiang and Liu, Yong},
  booktitle=AAAI,
  year={2025}
}

@article{drivearena,
    title={DriveArena: A Closed-loop Generative Simulation Platform for Autonomous Driving}, 
    author={Xuemeng Yang and Licheng Wen and Yukai Ma and Jianbiao Mei and Xin Li and Tiantian Wei and Wenjie Lei and Daocheng Fu and Pinlong Cai and Min Dou and Botian Shi and Liang He and Yong Liu and Yu Qiao},
    journal={arXiv preprint arXiv:2408.00415},
    year={2024}
}

@article{gpt4,
  title={Gpt-4 technical report},
  author={Achiam, Josh and Adler, Steven and Agarwal, Sandhini and Ahmad, Lama and Akkaya, Ilge and Aleman, Florencia Leoni and Almeida, Diogo and Altenschmidt, Janko and Altman, Sam and Anadkat, Shyamal and others},
  journal={arXiv preprint arXiv:2303.08774},
  year={2023}
}

@article{zhu2023minigpt,
  title={Minigpt-4: Enhancing vision-language understanding with advanced large language models},
  author={Zhu, Deyao and Chen, Jun and Shen, Xiaoqian and Li, Xiang and Elhoseiny, Mohamed},
  journal={arXiv preprint arXiv:2304.10592},
  year={2023}
}

@inproceedings{liu2023llava,
  title={Visual instruction tuning},
  author={Liu, Haotian and Li, Chunyuan and Wu, Qingyang and Lee, Yong Jae},
  booktitle=NeurIPS,
  year={2024}
}

@inproceedings{nie2023reason2drive,
  title={Reason2drive: Towards interpretable and chain-based reasoning for autonomous driving},
  author={Nie, Ming and Peng, Renyuan and Wang, Chunwei and Cai, Xinyue and Han, Jianhua and Xu, Hang and Zhang, Li},
  booktitle=ECCV,
  year={2024}
}

@inproceedings{qian2024nuscenes,
  title={Nuscenes-qa: A multi-modal visual question answering benchmark for autonomous driving scenario},
  author={Qian, Tianwen and Chen, Jingjing and Zhuo, Linhai and Jiao, Yang and Jiang, Yu-Gang},
  booktitle=AAAI,
  year={2024}
}

@inproceedings{sima2023drivelm,
  title={Drivelm: Driving with graph visual question answering},
  author={Sima, Chonghao and Renz, Katrin and Chitta, Kashyap and Chen, Li and Zhang, Hanxue and Xie, Chengen and Luo, Ping and Geiger, Andreas and Li, Hongyang},
  booktitle=ECCV,
  year={2024}
}

@article{mao2023gptdriver,
  title={Gpt-driver: Learning to drive with gpt},
  author={Mao, Jiageng and Qian, Yuxi and Ye, Junjie and Zhao, Hang and Wang, Yue},
  journal={arXiv preprint arXiv:2310.01415},
  year={2023}
}

@inproceedings{wen2023dilu,
  title={Dilu: A knowledge-driven approach to autonomous driving with large language models},
  author={Wen, Licheng and Fu, Daocheng and Li, Xin and Cai, Xinyu and Ma, Tao and Cai, Pinlong and Dou, Min and Shi, Botian and He, Liang and Qiao, Yu},
  booktitle=ICLR,
  year={2024}
}

@article{huang2024drivemm,
  title={Drivemm: All-in-one large multimodal model for autonomous driving},
  author={Huang, Zhijian and Feng, Chengjian and Yan, Feng and Xiao, Baihui and Jie, Zequn and Zhong, Yujie and Liang, Xiaodan and Ma, Lin},
  journal={arXiv preprint arXiv:2412.07689},
  year={2024}
}

@article{openvla,
    title={OpenVLA: An Open-Source Vision-Language-Action Model},
    author={{Moo Jin} Kim and Karl Pertsch and Siddharth Karamcheti and Ted Xiao and Ashwin Balakrishna and Suraj Nair and Rafael Rafailov and Ethan Foster and Grace Lam and Pannag Sanketi and Quan Vuong and Thomas Kollar and Benjamin Burchfiel and Russ Tedrake and Dorsa Sadigh and Sergey Levine and Percy Liang and Chelsea Finn},
    journal = {arXiv preprint arXiv:2406.09246},
    year={2024},
}

@inproceedings{neuroncap,
  title={Neuroncap: Photorealistic closed-loop safety testing for autonomous driving},
  author={Ljungbergh, William and Tonderski, Adam and Johnander, Joakim and Caesar, Holger and {\AA}str{\"o}m, Kalle and Felsberg, Michael and Petersson, Christoffer},
  booktitle=ECCV,
  year={2024},
}

@inproceedings{momAD,
  title={Don't Shake the Wheel: Momentum-Aware Planning in End-to-End Autonomous Driving},
  author={Song, Ziying and Jia, Caiyan and Liu, Lin and Pan, Hongyu and Zhang, Yongchang and Wang, Junming and Zhang, Xingyu and Xu, Shaoqing and Yang, Lei and Luo, Yadan},
  booktitle=CVPR,
  year={2025}
}

@article{hu2024drivingworld,
  title={DrivingWorld: Constructing world model for autonomous driving via video GPT},
  author={Hu, Xiaotao and Yin, Wei and Jia, Mingkai and Deng, Junyuan and Guo, Xiaoyang and Zhang, Qian and Long, Xiaoxiao and Tan, Ping},
  journal={arXiv preprint arXiv:2412.19505},
  year={2024}
}

@misc{openscene2023,
      title = {OpenScene: The Largest Up-to-Date 3D Occupancy Prediction Benchmark in Autonomous Driving},
      author = {OpenScene Contributors},
      howpublished={\url{https://github.com/OpenDriveLab/OpenScene}},
      year = {2023}
}

@misc{liu2024llavanext,
    title={LLaVA-NeXT: Improved reasoning, OCR, and world knowledge},
    url={https://llava-vl.github.io/blog/2024-01-30-llava-next/},
    author={Liu, Haotian and Li, Chunyuan and Li, Yuheng and Li, Bo and Zhang, Yuanhan and Shen, Sheng and Lee, Yong Jae},
    year={2024}
}

@article{zhang2025epona,
  title={Epona: Autoregressive Diffusion World Model for Autonomous Driving},
  author={Zhang, Kaiwen and Tang, Zhenyu and Hu, Xiaotao and Pan, Xingang and Guo, Xiaoyang and Liu, Yuan and Huang, Jingwei and Yuan, Li and Zhang, Qian and Long, Xiao-Xiao and others},
  journal={arXiv preprint arXiv:2506.24113},
  year={2025}
}

@inproceedings{para-drive,
  title={Para-drive: Parallelized architecture for real-time autonomous driving},
  author={Weng, Xinshuo and Ivanovic, Boris and Wang, Yan and Wang, Yue and Pavone, Marco},
  booktitle=CVPR,
  year={2024}
}

@inproceedings{Transfuser,
  author = {Prakash, Aditya and Chitta, Kashyap and Geiger, Andreas},
  title = {Multi-Modal Fusion Transformer for End-to-End Autonomous Driving},
  booktitle = CVPR,
  year = {2021}
}

@article{drama,
  title={Drama: An efficient end-to-end motion planner for autonomous driving with mamba},
  author={Yuan, Chengran and Zhang, Zhanqi and Sun, Jiawei and Sun, Shuo and Huang, Zefan and Lee, Christina Dao Wen and Li, Dongen and Han, Yuhang and Wong, Anthony and Tee, Keng Peng and others},
  journal={arXiv preprint},
  year={2024}
}

@article{drivinggpt,
  title={Drivinggpt: Unifying driving world modeling and planning with multi-modal autoregressive transformers},
  author={Chen, Yuntao and Wang, Yuqi and Zhang, Zhaoxiang},
  journal={arXiv preprint},
  year={2024}
}

@article{world4drive,
  title={World4Drive: End-to-End Autonomous Driving via Intention-aware Physical Latent World Model},
  author={Zheng, Yupeng and Yang, Pengxuan and Xing, Zebin and Zhang, Qichao and Zheng, Yuhang and Gao, Yinfeng and Li, Pengfei and Zhang, Teng and Xia, Zhongpu and Jia, Peng and others},
  journal={arXiv preprint},
  year={2025}
}

@misc{ImpromptuVLA,
      title={Impromptu VLA: Open Weights and Open Data for Driving Vision-Language-Action Models}, 
      author={Haohan Chi and Huan-ang Gao and Ziming Liu and Jianing Liu and Chenyu Liu and Jinwei Li and Kaisen Yang and Yangcheng Yu and Zeda Wang and Wenyi Li and Leichen Wang and Xingtao Hu and Hao Sun and Hang Zhao and Hao Zhao},
      year={2025},
      eprint={2505.23757},
      archivePrefix={arXiv},
      primaryClass={cs.CV},
      url={https://arxiv.org/abs/2505.23757}, 
}
}



\addtolength{\textheight}{-12cm}   


\end{document}